\documentclass{article}

\usepackage{arxiv}

\usepackage[utf8]{inputenc} 
\usepackage[T1]{fontenc}    
\usepackage{hyperref}       
\usepackage{url}            
\usepackage{booktabs}       
\usepackage{amsfonts}       
\usepackage{nicefrac}       
\usepackage{microtype}      
\usepackage{lipsum}

\title{An original framework for Wheat Head Detection using Deep, Semi-supervised and Ensemble Learning within Global Wheat Head Detection (GWHD) Dataset}

\usepackage{authblk}
\author[1]{\textbf{Fares Fourati}  }
\author[1, 2]{\textbf{Wided Souidene} }
\author[1]{\textbf{Rabah Attia} }
\affil[1]{ SERCOM Laboratory, Ecole Polytechnique de Tunisie, University of Carthage, Tunis, Tunisia; }
\affil[2]{ L2TI, Institut Galilée , Université Sorbonne Paris Nord, Villetaneuse, France }

\begin{document}
\maketitle

\begin{abstract}
In this paper, we propose an original object detection methodology applied to Global Wheat Head Detection (GWHD) Dataset. We have been through two major architectures of object detection which are FasterRCNN and EfficientDet, in order to design a novel and robust wheat head detection model. We emphasize on optimizing the performance of our proposed final architectures. Furthermore, we have been through an extensive exploratory data analysis and adapted best data augmentation techniques to our context. We use semi supervised learning to boost previous supervised models of object detection. Moreover, we put much effort on “ensemble” to achieve higher performance. Finally we use specific post-processing techniques to optimize our wheat head detection results. Our results have been submitted to solve a research challenge launched on the GWHD Dataset  which is led by nine research institutes from seven countries. Our proposed method was ranked within the top 6\% in the above mentioned challenge.

\end{abstract}


\section{Introduction}
Precise wheat head detection in outdoor field images can be visually difficult. Actually, there are generally  two main issues: overlap of dense wheat plants and the blur effect of the wind on the photographs. Both issues make it hard to detect single heads. Furthermore, wheat head appearance differs due to maturity, color, genotype, and head orientation. Finally, as wheat grows all over the world, different varieties, planting densities, patterns, and field conditions must be found. Thus, models developed for wheat head detection need to be robust for different growing environments. Nowadays, a bias due to the training region is present even for methods trained with a large data set. 
The aim of this paper is to design a robust object detection model, able to identify wheat heads in an image containing plenty of wheat heads and generalize on all types of wheat from all over the world. So, we propose a model that can accurately detect wheat heads even for new and unseen wheat types from a region or a country that are not included in the training dataset. To ensure such performance we design a novel architecture and use, for the very first time in the literature, a new dataset that contains images from different countries. We will use images taken from Europe (France, United Kingdom, Switzerland) and North America (Canada) for the training of our models and images from Australia, Japan, and China for the testing. So, we had the opportunity to train our models on diverse wheat images which helped us use more robust train, validation and test pipeline to assess the robustness of our algorithms. To optimize the performance of our models we combine different techniques such as pseudo labeling, test time augmentations, multi-scale ensemble, bagging and different post processing algorithms. By that we showed the added value of each component and designed a full architecture remarkably achieving higher performance.

\textbf{Contributions:}
\begin{itemize}
\item This paper uses a novel solution proposed recently for Object Detection using Deep Learnig: EfficientDet \cite{tan2020efficientdet}
\item This is the very first time, the data set GWHD is used in a research paper.
\item The proposed framework is unique and includes innovative methods based on data transformation and various model regularization techniques.
\item The performances obtained with our framework were submitted to solve a research challenge launched on the GWHD Dataset \url{http://www.global-wheat.com/} \cite{david2020global} and was ranked within the top 6\%.
\end{itemize}

The rest of this paper is organized as follows. In section 2, we propose an overview of related works on deep learning object detection architectures in general and in wheat head detection in particular. In section 3, we will dive into the proposed approach, we will expound techniques and ideas we incorporated and developed for data transformation, deep model regularization, semi-supervised learning, ensemble and post-processing to achieve optimal precision. Before concluding, we depict, in section 4, the experimental results.

\section{Related Work}


In this section we will deal with deep learning methods used for object detection and then we will explain some proposed research works for wheat head detection. Deep learning for object detection has been widely used since few years. The algorithms and approaches proposed in this field are divided into two genres: “two-stage detection” and “one-stage detection”.

\subsection{Two-stage detection}
 One of the most famous network within this family is the R-CNN (Region-based Convolutional Neural Network) which is based on the following steps: firstly, it extracts a set of object proposals by selective search \cite{uijlings2013selective}, then every proposal is re-sized and fed to a convolution neural network algorithm to extract parameters. In the end, SVM (Support Vector Machine) classifiers are implemented to decide the existence of an object in every region and to identify object classes. Fast R-CNN detector \cite{girshick2015fast} was published as an evolution of R-CNN. Fast R-CNN brings a design that simultaneously trains a classifier and regressor under the same network configurations. This achieved a speed over 200 times faster than R-CNN. After that, Faster-RCNN \cite{ren2015faster} introduced a novel approach called the Region Proposal Network (RPN) that achieved almost cost-free region proposals.  

\subsection{One-Stage Detection}
Huge advancements have been made recently towards more precise object detection; at the same time, state of the art object detectors also become more and more demanding. To tackle this problem, Google Research Brain Team developed a novel object detection architecture. In their paper \cite{tan2020efficientdet}, they studied several neural network architecture options and showed different optimization keys for efficiency improvement. They proposed a weighted bi-directional feature pyramid network (BiFPN), that ensures easier and faster multi-scale fusion of features. Also, they proposed a blended scaling technique that scales the resolution, depth, and width uniformly for all backbone, feature network, and box prediction neural networks at the same time. Using these optimizations and EfficientNet \cite{tan2019efficientnet} backbones, they have developed a novel family of object detection algorithms, called EfficientDet, that in a fair and impartial way get much better efficiency than previous architectures. Specifically, EfficientDetD7 achieves SoTA by being around 4x to 9x smaller and using 13x to 42x fewer FLOPs than previous detectors.\cite{tan2020efficientdet}

In the specific field of Wheat Head Detection, a study of the state of the art shows that one of the limitations is that proposed solutions are developed for controlled environment images and not images captured directly in the field \cite{pound2017deep}. Another limitation is the focus on the same wheat type for the train and the test of the models which leads to overfitting to that specific wheat type \cite{madec2019ear}. Another limitation is the focus on $mAP_{0,5}$ metric. In other words, a wheat head detected with a 50\% of Intersection over Union (IoU) is considered as a right detection. In this paper, to solve the first limitation, we use the GWHD dataset \cite{david2020global} that contains images captured in the field and shows more diversity in terms of sources, acquisition, conditions and wheat types. And to solve the second limitation we will use a different metric: $mAP_{0.5:0.75}$. In addition, the high occurrence of overlapping and occluded objects is unique in the GWHD Dataset. Moreover, we put much effort on data transformation and deep models' regularization to achieve a generalized solution for all wheat images. The GWHD dataset is published in May 2020. Although we are not the only ones to work on this dataset, we did not find published research papers yet. Therefore, we have used two different general object detection algorithms and designed better architectures based on those algorithms. In the results section we will consider our first models as references and study the impact of our appended ideas.

\section{Proposed Approach}

Our approach to tackle the problem of wheat head detection in outdoor field images consists of four major steps. First step is the Exploratory Data Analysis (EDA) which we included in our architecture in order to exploit our data in the best way. Second step is the semi-supervised learning. Third, is the ensemble and finally a post-processing block to enhance the accuracy of our whole architecture. Each step will be detailed in the following subsections.

\subsection{Data Transformation}
Before describing our Data Transformation strategy, it’s worth to mention that in deep learning, exploratory data analysis is a mandatory part of the development process, it’s not just an element of experimental set-up. Exploratory data analysis had a mandatory part of our work and it helped us achieve optimal inference results. It is a continuous process. We analyze data, formulate hypotheses, build models, make inferences, validate, compare and return back to analyzing the data until we achieve satisfying results.
 Because the GWHD dataset is the fusion of 9 sub-datasets, we have studied the characteristics of each to understand the specificity of each one of them. We have been through the details of each source including the differences in the wheat types and the image acquisition processes. In addition, we carefully studied the labeling quality and tried to clean some mistakes. Moreover, we used specific techniques to split the public data to provide a local training and validation pipeline. Furthermore, in this paper, we implement different data augmentation techniques to augment the size of our dataset, to provide more possible variations, to minimize overfitting and to guarantee the generalization of our models. In our choices of data augmentations methods, we relayed on characteristics of wheat images, geometrical constraints, intuition and experiments. 
Along with our own analysis of the images, we have been through the paper describing the GWHD dataset \cite{david2020global}  to get further details and check the provided meta-data.

\subsubsection{Data Analysis}

We used 3422 images from Europe (France, UK, Switzerland) and North America (Canada) for the training and the validation. And about 1,000 images from Australia, Japan, and China to test our models. In the 3422 images we found 49 images without wheat heads labels. We have neglected these images and we only used the labeled ones. So, we left with 3373 labeled images. In the 3373 images we found around 147793 bounding boxes. \\
We studied the sizes of bounding boxes and we found out that most wheat heads have small sizes. In addition, we found that most of the images have a covered area of bounding boxes of 20 to 40\%. We also found images with very high and others with very low brightness. Obviously, the images are highly diverse as they come from different countries, regions and sources. Those differences are mainly due to the difference of wheat genotypes, wheat agriculture conditions and the image acquisition diversity.

\subsubsection{Data Cleaning}

Data cleaning is the process of identifying incomplete, incorrect, inaccurate or irrelevant parts of the data and then replacing, modifying, or deleting them. Data cleaning is a mandatory part to prevent the misleading of our models and to improve the precision of our results. When analyzing the bounding boxes, we found huge bounding boxes covering more than one wheat head, and we found tiny bounding boxes smaller   than any wheat head. Probably that is due to the initial labeling process. So, we have deleted the huge and tiny bounding boxes by putting some threshold considering the range of the expected sizes of wheat heads. 

\subsubsection{Data Splitting}
To have trustworthy validation precision, it is mandatory to have a reasonable validation dataset. By splitting the data randomly, we could get by chance a non-equitable distribution of classes or types or a certain criteria between the train and validation. That could lead to non-reliable validation scores. This is especially the case when the data is imbalanced for a certain criterion. As images come from different sources (First Criteria), and images don't have approximately the same number of wheat heads per image. Some images are dense whereas others are almost empty (Second Criteria). Therefore, we used stratified k folds splitting. We split the data into 5 folds and made sure that each fold has approximately the same distribution of bounding boxes per image and the same number of images per source. In that way our cross validation was more reliable than just splitting randomly.

\subsubsection{Data Augmentation}
In recent years, a lot of effort have been invested in the development of novel algorithms and approaches. While artificial neural network architectures have been investigated in depth, less focus has been put into designing strong types of data augmentation that capture better patterns. During our experiments we have seen the greater impact of varying data augmentation methods on improving the mAP of our models. Which made us invest enough time on this part to optimize our results. Although most previous neural networks approaches used basic data augmentation types like flipping, rotations, padding and cropping, some research groups tried to design novel types of data augmentations for the task of classification \cite{shorten2019survey}. We adapted and inspired other strong augmentation methods for the task of object detection.  Most deep learning algorithms need huge quantities of data to achieve interesting results. Although most images are dense with wheat heads, around 3000 images could not be enough to train and validate our models. Which makes augmenting the number of images an important step. In fact, data augmentation could improve the generalization and reduce the over-fitting of a model by making different possible variations of the same image. This smart technique significantly increases the diversity of the available data for training, which makes the model learn diverse situations and thus, makes it more robust to various inputs. Although, we believe that data augmentation is mandatory to achieve the highest scores, we are aware that the usage of these methods should be with caution. Some data augmentations could make the task ambiguous. The augmentation techniques that we used: horizontal and vertical flips, rotations, cropping and resizing, random erasing \cite{zhong2020random}, cutout augmentation \cite{devries2017improved}, Cutmix augmentation \cite{yun2019cutmix}, gaussian noise injection \cite{moreno2018forward}, and natural noise injection such as simulating the wind effect with a moving blur augmentation and simulate shadow due to the acquisition process and the position of sun using shadow augmentation \cite{buslaev2020albumentations}. We tried to adapt all those techniques to the context of wheat head detection.

\subsection{Semi-supervised learning: Pseudo Labeling (PL)}
We have used a pseudo labeling \cite{lee2013pseudo} as our semi-supervised algorithm during the inference time. Because we are not building a real time object detection application, trying pseudo labeling was a considerable option. It works as follows: in the first step we train the chosen object detector in a supervised manner, then that trained model will predict labels for the test set. Then, our model will consider those pseudo-labels as if they are the ground truth and append them to the training data set. Then, our model re-trains on the new training data set that includes the labels and the pseudo labels for a certain number of epochs. Then, our model will re-predict new bounding boxes for the test set. We could repeat these steps for N rounds.

\subsection{Ensemble}
\subsubsection{Test Time Augmentation (TTA) }
While data augmentation is done before or while the training of the model, this one is done during the inference time. It is a simple but yet a very effective way to optimize mAP results. In the context of object detection, test time augmentation could lead to more accurate results in terms of detecting more wheat heads. We have relied on this technique on our architectures and have shown its added value. Although its great impact on the precision of deep models, only few papers discussed this method. The idea is to show different versions of the same image to the same model, take the different outputs and extract the detected wheat heads and then combine the results using some post processing techniques. While test time augmentation, like training data augmentation, could be simple to implement and use for classification, it would lead to difficulties in object detection due to the complexity of the task. So the use of this method should be with caution. Therefore, TTA is more common in classification than object detection. It is mandatory to use reversible transforms because after inference we need to transform predictions on the original image. We have compared different transforms including varying brightness or contrast but later we have only focused on test time augmentation based on orientation transformations including vertical flips, horizontal flips and 90 degrees rotation. From one image we produce $2^{3}$ versions and get predictions of each version of these 8 images and then combine the results.

\subsubsection{Bootstrap Aggregating (Bagging)}
Bootstrap aggregating \cite{breiman1996bagging}, also called bagging, is a machine learning ensemble algorithm designed to increase the stability and the precision of ML algorithms. It also reduces variance and helps to avoid overfitting.
The idea behind bagging is combining the results of multiple models to get a generalized result. If these models are trained over the exact data, they are more likely to predict almost the same results. To overcome this point, bagging consists in training models over different subsets of data to get more diversity during the inference time which leads to better results after the combination of those inferences.

\subsubsection{Multi-scale ensemble}
We have trained different EfficientDet models over different input sizes, specifically 512x512 and 1024x1024. Although bigger images, intuitively could provide higher accuracy, they are very expensive to use for the training. Which explains our preference in the use of 512x512 images for our experiments. However, we make sure to train a few models on 1024x1024 input resolution to compare the results and most importantly to ensemble both approaches.

\subsubsection{Post-processing}

Object detection models predict more bounding boxes than wheat heads in the image. Therefore, post-processing algorithms are necessary to finally bring the right bounding boxes. Because we used ensemble, post-processing was mandatory to combine the various predictions to come up with a better final prediction.
A classical and yet very effective way to do post processing is by using Non-Maximum-Suppression (NMS). It tries to detect the best prediction among multiples overlapping predictions. However, we believe that Weighted Boxes Fusion (WBF) \cite{solovyev2019weighted} is a better choice for ensemble. In contrast to Non-maximum-suppression that chooses best predictions and deletes the others, WBF computes a weighted average of overlapped predictions to bring a more accurate prediction. So, WBF extracts information from all the predictions rather than choosing the best. That follows the same spirit of ensemble where we need to take information from different predictions.

\section{Experiments}
In this section, our experimental tests are presented. First, the tests platform will be presented. Then, we are going to highlight some results and test performances.

\subsection{Faster-RCNN}
We have started from the pre-trained model on pedestrian images and done some fine-tuning. We only fine-tuned the models for a few epochs. Used a Resnet50 backbone, Stochastic Gradient Descent as an optimizer, a simple learning rate scheduler that decreases the LR after a certain number of epochs (steps), momentum 0.9, weight decay 0.0005. Compared the use of cleaning, local data augmentations and pseudo labeling. Finally, we studied the impact of each of these techniques on the mAP of our models. (Tab. 1)

\begin{table}[h!]
\centering
{
 \begin{tabular}{||c c c c c c||} 
 \hline
 Image Size & Batch & Data Cleaning & Data Augmentation & Psuedo Labeling & mAP\% \\ 
 \hline\hline
 512 & 16 &  - & - & - &63.09 \\ 
  512& 16 &  True & x2 & - &64.98 \\
 512& 16 &  - & - & True &65.84 \\
 512& 16 &  True & - & True &66.92 \\
 512& 16 &  True & x4 & - &67.24 \\
 512& 16 &  True & x4 & True &\textbf{68.46} \\
 \hline
 \end{tabular}}
 \caption{Performance of the data transformation techniques and pseudo labeling on the framework overall mAP.}
\label{table:1}
\end{table}

\subsection{EfficientDet}

For EfficientDet training, AdamW was our optimizer. We reduced the batch size to 4 and used different learning rate scheduler than the one we used for Faster RCNN which reduces the learning rate only when achieving a plateau for certain number ’patience’ of epochs. Rather than using local augmentations we used online method and we used weighted boxes fusion for post-processing. The first model was based on EfficientDet5 architectures. The training has been based on online data augmentation. So, each augmentation has its own probability of realization. We have chosen the intervals of those probabilities with intuition and experiments by following the training and validation losses. In this paper we will limit our augmentation comparison to two groups G1 and G2 (Tab. 2) where we apply on each image each transformation, during the training of our models, with a certain probability. We studied the impact of the previously discussed techniques on the mAP. (Tab. 3)

\begin{table}[h!]
\centering
{ 
\begin{tabular}{||c c c ||} 
 \hline
 Image Transformations & G1 Probabilities & G2 Probabilities \\ 
 \hline\hline
  Crop and Resize & 0.5 & 0.2\\
 Hue Saturation Value & 0.8 & 0.8 \\
 Brightness Contrast &   - & 0.8 \\ 
 To Gray &  0.01 &  0.01\\ 
 Inject Gaussian Noise &  - &0.01 \\ 
  Horizontal Flip&  0.5 & 0.4 \\ 
 Vertical Flip & 0.5 & 0.4  \\ 
  Random Rotate 90 & - & 0.4 \\ 
Cutout (8 holes) & 0.5 & 0.4 \\ 
Cutout (10 holes) &- & 0.4 \\ 
Motion Blur (Wind) &- & 0.3 \\ 
Shadow &- & 0.3 \\ 
 \hline
 \end{tabular}}\\
 \caption{During the training of our models we apply on each image each transformation with a certain probability.}
\label{table:2}
\end{table}

\begin{table}[h!]
\centering
{ 
\begin{tabular}{||c c c c c c c||} 
 \hline
 Image Size & Data Augmentation & TTA & Pseudo Labeling & Bagging & Multi Scale                       & mAP \% \\ 
 \hline\hline
  512 & G1 & - & - & - & - &71.23 \\
 512& G1 & -&-&- &-&71.93 \\ 
 512&   G2 & -&-& -&-&72.11 \\ 
 512&  G2 & -& True & -&-&72.33 \\ 
 512&  G2 & True &  -& -&-&73.29 \\ 
  512&  G2 & True & True & -&-&73.36 \\ 
 512& G2 & - & True & True &-&73.43 \\ 
  512 \& 1024 &  G2 & True & - & True &True&\textbf{74.22} \\ 
 \hline
 \end{tabular}}
 \caption{ Performance of the whole framework for different designed steps.}
\label{table:3}
\end{table}

\section{Conclusion}
In this paper, we designed an original framework for wheat head detection within a newly released dataset (GWHD). We mainly focused on regularization of deep model methods. We investigated ensemble and semi-supervised learning methods to push the precision of our models. We have been through the data, analyzed the sources, types, distribution, diversity, specificity and used this knowledge to improve our training processes. Basically, we have designed custom data augmentation techniques to the wheat dataset. We have developed a whole object detection framework and adapted it to the context of wheat head detection. We’ve finally achieved a robust object detector and we’ve been ranked in the top 6\% in the Wheat Head Detection challenge.

\bibliographystyle{unsrt}  
\bibliography{references}  



\end{document}